\pdfoutput=1

\documentclass[11pt]{article}

\usepackage{ACL2023}

\usepackage{times}
\usepackage{latexsym}

\usepackage[T1]{fontenc}

\usepackage[utf8]{inputenc}

\usepackage{microtype}

\usepackage{inconsolata}
\usepackage{booktabs}
\usepackage{multirow}
\usepackage{dingbat}
\usepackage{float}
\usepackage{amsmath}
\usepackage{amssymb}
\usepackage{times}
\usepackage{latexsym}
\usepackage{microtype}
\usepackage{color, colortbl}

\usepackage[textsize=scriptsize]{todonotes}

\usepackage{graphicx}
\graphicspath{ {./} }

\title{Drafting Event Schemas using Language Models}

\author{Anisha Gunjal \quad\quad\quad Greg Durrett \\
        Department of Computer Science \\
        The University of Texas at Austin \\
        \texttt{anishagunjal@utexas.edu}
}

\begin{document}
\maketitle

\begin{abstract}
Past work has studied event prediction and event language modeling, sometimes mediated through structured representations of knowledge in the form of event schemas. Such schemas can lead to explainable predictions and forecasting of unseen events given incomplete information. In this work, we look at the process of creating such schemas to describe complex events. We use large language models (LLMs) to draft schemas directly in natural language, which can be further refined by human curators as necessary. Our focus is on whether we can achieve sufficient diversity and recall of key events and whether we can produce the schemas in a sufficiently descriptive style. We show that large language models are able to achieve moderate recall against schemas taken from two different datasets, with even better results when multiple prompts and multiple samples are combined. Moreover, we show that textual entailment methods can be used for both matching schemas to instances of events as well as evaluating overlap between gold and predicted schemas. Our method paves the way for easier distillation of event knowledge from large language model into schemas.


\end{abstract}

\section{Introduction}

\begin{figure}
    \centering
    \includegraphics[scale=0.58,trim=2mm 70mm 130mm 30mm]{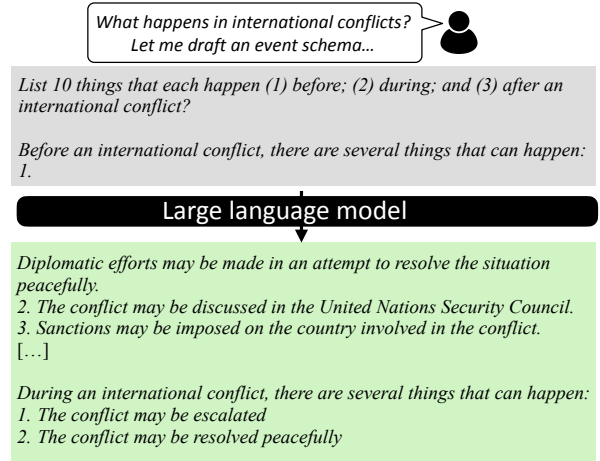}
    \caption{Overview of our system. A large language model can generate lightly structured lists of events, which themselves may have complex predicate-argument structure.}
    \label{fig:intro}
\end{figure}

Predicting and modeling sequences of events has become more sophisticated over the past decade. Early work mined narrative schemas that were limited in representational power: initially sequences of predicate-role pairs \cite{chambers-jurafsky-2008-unsupervised}, then generalized to predicate-argument structures \cite{chambers-jurafsky-2009-unsupervised}, which continue to be used in neural approaches \cite{weber-etal-2018,koupaee-etal-2021-dont}. Recently, language modeling provides a very flexible interface for predicting tokens given context and has been applied to event prediction \cite{rudinger-etal-2015-script,pichotta-mooney-2016-using,koupaee-etal-2021-dont} and cloze tasks \cite{paperno-etal-2016-lambada}. Event schemas can only compete with language modeling approaches if they are high-quality and specific enough to provide strong predictions. Ultimately, event schemas might enable explainable forecasting \cite{zou-etal-forecasting-2022} grounded in an expert-curated knowledge structure.

This paper attempts to bridge this gap by constructing natural language represents of event knowledge from language models. Past efforts like ATOMIC \cite{sap-etal-2019} and COMET \cite{bosselut-etal-2019-comet} show that structured repositories of knowledge and the ability to extend them can help enable predictions about the world. We follow in their vein and construct collections of events we call \emph{light schemas}. These are less structured than graph-based schemas anchored in event ontologies \cite{li-etal-2021-future}. Our chief aim is to have high recall over a set of events in a domain to serve as a draft for curation of a more structured schema.

We generate these schemas using language models like GPT-3.5 \cite{Brown-etal-2020-gpt3,ouyang2022training} and Flan-T5 \cite{chung-etal-2022-flant5}. As shown in Figure~\ref{fig:intro}, these models have strong abilities to surface events characteristic to a particular domain (e.g., \emph{international conflict}), including typical arguments for those events. Although our schemas are ontology-free, they implicitly have a certain ``style'' associated with the natural language expressions of their events. We explore both zero-shot and few-shot (specifically one-shot) prediction of schemas.

Understanding the event coverage of our schemas requires comparing them to schemas built by human curators. Evaluation of schematic knowledge and what it predicts have typically been restricted to cloze tasks \cite{Granroth-Wilding_Clark_2016,modi-etal-2017-modeling,weber-etal-2018} or events in certain coarse ontologies \cite{li-etal-2021-future}, but these do not directly evaluate schema representations themselves. Recent past work uses measures very tied to lexical expression of schema predicates \citep{dror2022zero,zhang2023humanintheloop}, but these are most appropriate for schemas in closed ontologies.

Instead, we evaluate our schema generation using textual entailment methods \cite{dagan-rte,williams-etal-2018-broad}, following a similar application of these methods to evaluate groundedness of summaries
\citep{falke-etal-2019-ranking,zhang2021finding,laban-etal-2022-summac}. We use entailment to compare our drafted schemas to two sources of ground-truth schemas annotated by human annotators. Specifically, we investigate whether an event we generate entails an event in the ground-truth schema as a measure of recall; we also explore bidirectional entailment (is there mutual entailment between the events?) for a more precise measure.

Through human study, we validate that our entailment-based evaluation is reliable. Our results show that large language models can generate schemas that have substantial overlap with ground-truth schemas written by curators. One-shot prediction allows us to emulate the stylistic features of target schemas and attain varying levels of specificity with respect to arguments of predicates. We compare different methods and find that drawing multiple samples from these models can further improve recall.


Our main contributions are (1) We analyze the performance of current text generation models (GPT-3.5 and Flan-T5) for the task of generating lightly organized event schemas in a completely training-free regime. (2) We show promising results of using textual entailment as a metric to automatically measure event coverage by generated schemas. (3) We show that one-shot prediction can be used to achieve stylistic control over schema generation, suggesting a way to adapt approaches for different desired output formats.

\section{Methods}
\label{sec:prelim}
\subsection{Preliminaries} Our schemas are anchored to domains $d$. An example of $d$ shown in Figure~\ref{fig:intro} is \emph{international conflict}; these may be broad topics or more specific scenarios like \emph{roadside bombing attack}. In this work, they are not anchored to specific entity participants.
We define a schema $S = (\mathbf{s}_1, \ldots, \mathbf{s}_n)$ as an ordered\footnote{We preserve the ordering of $S$ because we find that it often corresponds to a partial temporal ordering. We do not evaluate this aspect extensively in this work.} collection of sentences expressing events. The $\mathbf{s}_i$ are sentences expressing events at a moderate level of generality; they are typically short descriptions and do not involve specific named entities. However, we do not structurally constrain their form. We refer to the \textbf{style} of the schema as a collection of surface-level factors including the average length in words of the $\mathbf{s}_i$ and the specificity of the events.

We explore two classes of models in this work. First, \textbf{zero-shot models} have the form $P(S \mid v_c(d))$; they condition on a verbalization $v$ of domain $d$, parameterized by a strategy $c$. For example, the prompt in Figure~\ref{fig:intro} has $d=\mathrm{international\ conflict}$ and the verbalizer \emph{List 10 things that each happen (1) before; (2) during; and (3) after [$d$]...Before an [$d$], there are several things that can happen: 1.}. This verbalizer is designed to produce a certain pattern of temporal information; in this case, the answer from the model separates into events occurring before, during, and after the conflict. Other verbalizers we explore look at aspects like cause and effect; a full list of verbalizers is included in the Appendix \ref{sec:appendix:prompts}.

The verbalizers give us control over attributes $c$; however, they do not necessarily allow us to specify a target style for the schema. We find that each model has certain styles it tends to generate in across a range of verbalizers.

We also explore \textbf{one-shot models} $P(S \mid v(d), S_\mathrm{demo})$ that condition on a schema demonstration as well as a verbalizer of the domain. Note that $S_\mathrm{demo}$ is a hand-authored schema (or post-edited output of the model) coming from a separate domain $d'$. We give examples of the prompts we use in Appendix \ref{sec:appendix:prompts}.

\subsection{Models Considered}

\paragraph{GPT-3.5 \texttt{text-davinci-003}} We experiment with the most capable of the OpenAI GPT-3.5  models \cite{Brown-etal-2020-gpt3}. According to \cite{gpt35}, \texttt{text-davinci-003} is an instruction-tuned model \cite{ouyang2022training} using reinforcement learning on models of human preference judgments.

\paragraph{Flan-T5} We also experiment with \cite{chung-etal-2022-flant5}. We use the XXL variant, which is 11B parameters. This allows us to see what is achievable with a smaller instruction-tuned model that can be more easily and cheaply run.

We qualitatively observed that \texttt{flan-t5-xxl} does not perform well on temporally-aided complex prompt as described in \ref{sec:appendix:prompts}. Hence we simplify the prompt into three independent prompts:
\begin{enumerate}
    \item List events that occur \emph{before} ...
    \item List events that occur \emph{during} ...
    \item List events that occur \emph{after} ...
\end{enumerate}
The outputs generated are minimally post-processed if necessary to extract the events generated.

\paragraph{Older GPT-3 variants} We also tried using older variants of GPT-3 model \cite{Brown-etal-2020-gpt3} such as \texttt{text-davinci-base}, however the generations consisted of a lot of redundancies and required a lot of human curation to extract relevant information from the output, refer appendix \ref{sec:appendix:davinci-base}. For this reason, we exclude the base GPT-3 model \texttt{text-davinci-base} from our main results.


\paragraph{Inference hyperparameters} For all models, we decode using nucleus sampling \cite{holtzman-etal-2020-nucleus} with hyperparameters top-p=1.0 and temperature set to 0.7. We do not perform any model training and use off-the-shelf models for our analysis. The GPT-3.5 variants are accessed through OpenAI's API with estimate compute cost amounting to less than \$100. To run inference for \texttt{flan-t5-xxl} we host the model on a \texttt{p3.16xlarge} AWS instance.

\section{Evaluation via Textual Entailment}
\label{sec:entailment-eval}
Inspection of our schemas (see Figure~\ref{fig:intro}, Table~\ref{tab:examples-all}) shows that they are very high quality. As we are using state-of-the-art autoregressive Transformer models, the fluency, coherence of the sequence of events, and linguistic quality of each individual event are very high and do not need to be the focus of our evaluation. Instead, the main question is to what extent the events we have cover the important events in the target domain; they may fail to do so as a result of reporting biases in text. We can compare these to human-written schemas; however, because our schemas are in natural language, we will need a sophisticated comparison function in order to do so. Here, we turn to textual entailment.

Our evaluation focuses on comparing a predicted schema $\hat{S}$ with a ground-truth, human-annotated schema $S^*$. $S^*$ is considered to contain events that we want to see represented in predicted schemas. Note that $S^*$ may not be exhaustive; that is, an event not in $S^*$ may still be considered of high quality and relevant. Therefore, our evaluation will focus on recall.

We use textual entailment models of the form $E: (\mathbf{s}_1,\mathbf{s}_2) \rightarrow \mathbb{R}$ to judge whether two sentences are matching events. An entailment model computes a distribution over three classes \{\emph{entailment}, \emph{neutral}, \emph{contradiction}\}. We set $E$ to return the probability of $\mathbf{s}_1$ entailing $\mathbf{s}_2$, ignoring the distinction between neutral and contradiction. Intuitively, a sentence like \emph{protests break out in the city} should entail (and be entailed by) a sufficiently similar event like \emph{civil unrest in the capital}. While there may be minor variation in the argument (e.g., \emph{city} vs.~\emph{capital}), the notion of entailment still approximately captures the appropriate similarity for this task.

Our goal is to compare an event $\mathbf{s}$ to an entire schema $S$. The recall score $r$ for an event ${s} \in {S^*}$ is then given by
\begin{equation}
r(s, \hat{S}) = \max_{\hat{s} \in \hat{S}} (\max(E(s, \hat{s}),E(\hat{s}, s)))
\end{equation}
maxing over the events in the predicted schema. As the level of specificity between the gold events and predicted events can sway in either direction, we run the entailment model in both directions (e.g. gold event \emph{entails} predicted event or vice versa).

We consider two variants of this procedure: \textbf{any-directional entailment} where we use the entailment model as described above, and \textbf{bidirectional entailment} where we modify the score to be $\min\{ E(s, \hat{s}), E(\hat{s}, s) \}$. This places a stronger requirement that the two statements be equivalent.

\paragraph{Entailment Models Used}
We test our generated schemas using the textual entailment model \texttt{roberta-large-wanli} by \cite{liu2022wanli} trained on the WANLI dataset. This model uses RoBERTa-large \cite{liu2019roberta} architecture and has 345M parameters. 

\begin{table*}[t]
\renewcommand{\tabcolsep}{1.5mm}
    \centering
    \small
    \begin{tabular}{c|c|c|c||c||c}
    \toprule
       \multirow{2}{*}{{\textbf{Domain}}} & {\multirow{2}{*}{}\textbf{Gold Schema}} & {\textbf{davinci-003}} & {\textbf{flan-t5-xxl}} & \textbf{davinci-003} & \textbf{Dror et al.} \\
       & & zero-shot & zero-shot & one-shot & one-shot\\ \midrule

         {\textbf{Natural Disaster}} & \multirow{3}{*} 
        \textbf{\# Events} & 24.33 & 21.67 & 39.22 &  4.67\\
        & \textbf{RESIN} & 0.33$\pm$0.24 & 0.4$\pm$0.13 & 0.56$\pm$0.19 &  0.11$\pm$0.1\\ 
        & \textbf{CuratedSchemas} & 0.41$\pm$0.29 & 0.29$\pm$0.09 & 0.40$\pm$0.12 &  0.14$\pm$0.06\\  \midrule

        {\textbf{International Conflict}} & \multirow{3}{*} 
        \textbf{\# Events} & 29.67 & 25.67 & 44.67 & 5.33\\
        & \textbf{RESIN} & 0.44$\pm$0.08 & 0.6$\pm$0.12 & 0.46$\pm$0.15 & 0.07$\pm$0.07 \\ 
        & \textbf{CuratedSchemas} & 0.73$\pm$0.06 & 0.45$\pm$0.1 &  0.55$\pm$0.16 & 0.09$\pm$0.04\\  \midrule

        {\textbf{Mass Shooting}} & \multirow{2}{*} 
        \textbf{\# Events} & 15.67 & 22.66 & 29 & 4.67\\
        & \textbf{RESIN} & 0.23$\pm$0.07 & 0.45$\pm$0.47 & 0.57$\pm$0.19 & 0.21$\pm$0.07\\ 
        & \textbf{CuratedSchemas} & 0.27$\pm$0.04 & 0.41$\pm$0.14 & 0.53$\pm$0.16 & 0.25$\pm$0.06 \\  \midrule
        
        {\textbf{Disease Outbreak}} & \multirow{2}{*} \textbf{\# Events} & 27 & 23.67 & 23.33 & 5.33\\
        & \textbf{RESIN} & 0.46$\pm$0.15 & 0.4$\pm$0.13 & 0.38$\pm$0.07 & 0.15$\pm$0.03\\ 
        & \textbf{CuratedSchemas} & 0.37$\pm$.14 & 0.24$\pm$0.05 & 0.29$\pm$0.06 & 0.07$\pm$0.04 \\  \midrule

        {\textbf{Kidnapping}} & \multirow{2}{*} \textbf{\# Events} & 15 & 21 & 23.56 &  6.33\\
        & \textbf{RESIN} & 0.52$\pm$0.17 & 0.33$\pm$0.2 &  0.42$\pm$0.09 &  0.26$\pm$0.06\\ 
        & \textbf{CuratedSchemas} & 0.52$\pm$0.12 & 0.37$\pm$0.08 &  0.54$\pm$0.1 & 0.44$\pm$0.1\\  \midrule

        {\textbf{IED}} & \multirow{2}{*} 
        \textbf{\# Events} & 18 & 23.33 & 32.11  & 6\\
        & \textbf{RESIN} & 0.23$\pm$0.04 & 0.44$\pm$0.07 & 0.53$\pm$0.13 & .11$\pm$0.02\\ 
        & \textbf{CuratedSchemas} & 0.17$\pm$0.03 & 0.37$\pm$0.05 &  0.42$\pm$0.13  & 0.15$\pm$0.05 \\ 
        \midrule
        \midrule

        {\textbf{Average Across Domains}} & \multirow{2}{*} 
         \textbf{RESIN \& CuratedSchemas} & \textbf{0.39} & \textbf{0.3958} & \textbf{0.47} &   \textbf{0.1708} \\ 
        
    \bottomrule
    \end{tabular} \vspace{-2mm}
    \caption{Event recall of zero-shot and one-shot performance performance of different language models measured against human curated gold schemas from two datasets. We use any-directional entailment. One-shot results are substantially better for certain domains and lead to generation of more events. However, all systems are able to generate a substantial number of matching events across the domains of interest.}
    \label{tab:event-recall}
    \vspace{-3mm}
\end{table*}

\section{Experimental Setup}

\subsection{Gold Schema}

We conduct experiments on the gold schemas from two datasets: RESIN-11 \cite{du2022resin} and CuratedSchemas, described below. The domains included in our dataset are \emph{international conflict, natural disaster, IED attacks, disease outbreak, mass shooting, and kidnapping}. We sample these domains as they are available in both datasets and gives us the opportunity to test various interesting aspects of schema datasets such as varying coverage, and style of event descriptions. More details on both dataset can be found in Appendix~\ref{sec:dataset-stats}. For the published RESIN-11 Schema, we modify the event structure into a natural language sentence as described in Appendix \ref{sec:resin_schemas}. 

We also use a separate set of schemas we call the CuratedSchemas set. These schemas were annotated by ourselves and our collaborators independently of the RESIN schemas. Appendix~\ref{sec:curated_schemas} describes these.

\subsection{Language Model Schema Generation}
We predominatly test the schema drafting performance of GPT-3 variant \texttt{text-davinci-003} and Flan-T5 variant \texttt{flan-t5-xxl}. Each prompt is used to generate 3 generations. We report statistics and event recall on average of 3 generations in the \ref{sec:results} section.

We can also over-generate predictions using diverse prompts and achieve a higher event recall with the possibility of generating incorrect or redundant events. To experiment this, we craft 3 different prompts and sample 3 generations from each using the \texttt{text-davinci-003}. We call this approach \textbf{prompt union}. More details on prompt union can be found in the appendix \ref{sec:appendix:prompts}.

A key aspect of natural language event schemas is stylistic variations across datasets and language models that generate it. As discussed in \ref{sec:prelim}, we use one-shot prompts to guide the models to generate outputs similar to the target dataset style.

\begin{table*}[t]
\renewcommand{\tabcolsep}{0.8mm}
    \centering
    \small
    \begin{tabular}{c|  c|c|c}
    \toprule
       {\textbf{Domain}} & {\multirow{2}{*}{}\textbf{Gold}} & {\textbf{Single Prompt}} & {\textbf{Prompt Union}} \\ \midrule

         {\textbf{Natural Disaster}} & \multirow{3}{*} 
         \textbf{\# Events} & 24.33 & \cellcolor{red!20} 187\\ 
        & \textbf{RESIN} & 0.33 & \cellcolor{green!20} 0.86\\ 
        & \textbf{CuratedSchemas} & 0.41 & \cellcolor{green!20} 0.80\\  \midrule

        {\textbf{International Conflict }} & \multirow{3}{*} 
        \textbf{\# Events} & 29.67 & \cellcolor{red!20} 215 \\ 
        &\textbf{RESIN} & 0.44 &  \cellcolor{green!20} 0.6 \\ 
        & \textbf{CuratedSchemas} & 0.73 &  \cellcolor{green!20} 0.78  \\  \midrule

        {\textbf{Mass Shooting}} & \multirow{3}{*} 
        \textbf{\# Events} & 15.67 & \cellcolor{red!20} 154\\ 
        & \textbf{RESIN} & 0.23 &  \cellcolor{green!20} 0.62  \\ 
        & \textbf{CuratedSchemas} & 0.27 & \cellcolor{green!20} 0.76  \\  \midrule

        {\textbf{Disease Outbreak}} & \multirow{3}{*} 
        \textbf{\# Events} & 27 & \cellcolor{red!20} 203\\ 
        & \textbf{RESIN} & 0.46 &  \cellcolor{green!20} 0.76 \\ 
        & \textbf{CuratedSchemas} & 0.37 & \cellcolor{green!20} 0.52 \\  \midrule

        {\textbf{Kidnapping}} & \multirow{3}{*}
        \textbf{\# Events} & 15 & \cellcolor{red!20} 127\\ 
        & \textbf{RESIN} & 0.52 & \cellcolor{green!20} 0.91 \\ 
        & \textbf{CuratedSchemas} & 0.52 & \cellcolor{green!20} 0.83 \\  \midrule

        {\textbf{IED}} & \multirow{3}{*} 
        \textbf{\# Events} & 18 & \cellcolor{red!20} 134\\ 
        & \textbf{RESIN} & 0.17 & \cellcolor{green!20}0.70 \\ 
        & \textbf{CuratedSchemas} & 0.23 & \cellcolor{green!20}0.63  \\ 
        
    \bottomrule
    \end{tabular} \vspace{-2mm}
    \caption{Diverse and instructive prompts improve the coverage of schema generation. We compare the single prompt version used in \ref{tab:event-recall} against using a \textbf{prompt union} method which uses three prompts to over-generate events and improve recall. This result shows that the method can be potentially used to increase event coverage with gold schemas while compromising precision.}
    \label{tab:event-recall-multi}
    \vspace{-3mm}
\end{table*}

\begin{table*}[t]
    \centering
    \small
    \renewcommand{\tabcolsep}{1mm}
    \begin{tabular}{r|cc|cc|c}
    \toprule
      
       \multirow{2}{*}{\textbf{Domain}}& RESIN-11 & CuratedSchemas & text-davinci-003 &  flan-t5-xxl & text-davinci-003 \\ 
       & & & zero-shot & zero-shot & one-shot \\ \midrule
       Natural Disaster & 4.47 & 3.36 & 5.68 & 4.75 & 6.68 \\
       International Conflict & 5.27 & 2.55 & 2.86 & 4.81 & 4.89 \\
       Mass Shooting & 5.24 & 3.32 & 8.6 & 6.62 & 6.02 \\ 
       Disease Outbreak & 7.57 & 4.45 & 5.08 & 7.1 & 5.95 \\
       Kidnapping & 7.08 & 4.5 & 5.07 & 6.67 & 6.11 \\
       IED & 4.87 & 3.24 & 5.48 & 6.87 & 8.11  \\ 
       \midrule
       Mean & 5.75 & 3.57 & 5.46 & 6.14 & 6.29  \\
    \bottomrule
    \end{tabular} \vspace{-2mm}
    \caption{Average length in words of the gold/generated events in each schema. We argue that word length is a proxy for a schema's style (\emph{political unrest} vs.~\emph{protestors cause civil unrest in the capital} differ in specificity in a way that length reveals).} 
    \label{tab:word-length}
    \vspace{-1em}
\end{table*}

\section{Results}
\label{sec:results}

\subsection{Schema Generation Performance}

Table \ref{tab:event-recall} shows the results of several schema generation approaches measured against the gold schemas from RESIN-11 and CuratedSchemas. The metric used to measure the recall is the \emph{any-directional entailment} as described in Section~\ref{sec:entailment-eval}. We report the mean and standard deviation of event recall across three sampled generations for each prompt. Along with the event recall, we also report the number of events predicted by each model, which gives signal about the precision of their event generation.

\paragraph{Zero-shot generation performance is high.} Table \ref{tab:event-recall} highlights that both \texttt{text-davinci-003} and \texttt{flan-t5-xxl} show an average of 0.39 and 0.3958 event coverage with respect to the gold schemas. Discussions on human agreement of the entailment model judgments and influence of generation style are deferred to Sections \ref{sec:human_eval_entailment} and \ref{sec:style}. We also report the average number of event generated by both models for each prompt which are in the range of 15-30, indicating that we do not over-generate events for each domain to increase the recall performance. Finally, the overlap of generated events with both the human curated gold schemas (RESIN-11 and CuratedSchemas) is substantial and reflects on the potential of language models in drafting complex schemas with sufficient coverage. 


\paragraph{Drawing more samples from this model can increase recall further.} Complex event schemas of domains like disease outbreak or natural disasters can have varying actors and topics that cannot be exhaustively sampled from a single prompt. For instance, \emph{What happens after a disease outbreak?} can have various responses talking about either \emph{legal proceedings against organisations who are held accountable for a disease outbreak} OR \emph{research on preventing future outbreaks of the disease} - both responses are valid but cover various aspects of the complex event. This result can be emulated by using diversity in prompts generation to generate events affecting different participants from the event. In Table~\ref{tab:event-recall-multi} we compare event coverage results from a single prompt versus taking a union across a larger number of prompts and samples. We see that taking a union of generations from various prompts leads to a substantial boost to the event recall with the caveat that we generate larger number of events.

\paragraph{Prompting models with complex prompts leads to significantly higher performance than past work}
\citet{dror2022zero} explore using language models to generate documents which can be used to construct complex schemas for events. Their work studies generation of direct step-by-step schemas using prompts such as \emph{What are the steps involved in topic? 1.}. We generate responses from this prompt template to extract natural language events. Our work is not a direct comparison to \cite{dror2022zero} as their focus is predominantly using language models to generate documents for downstream event schema induction pipelines. However, we only adopt their direct step-by-step schema generation prompt and argue that using complex prompts can lead to better event coverage in comparison to simpler prompts such as listing steps in an event. This result is highlighted by comparing the results of \texttt{text-davinci-003} and \texttt{Dror et al.} in Table~\ref{tab:event-recall}.

\begin{table}
\centering
\begin{tabular}{ll}
\toprule
\textbf{Gold Schema} & \textbf{Avg. Event Recall}\\
\midrule
RESIN & 0.62 \\
CuratedSchemas & 0.46  \\
\bottomrule
\end{tabular}
\caption{Measuring the overlap between events in the gold schemas: RESIN and CuratedSchemas. We use any-directional entailment to get an estimate of the overlap between two distinct human-curated schemas. }
\label{tab:schema-overlap}
\end{table}

\begin{table}[t!]
\centering
\small
\renewcommand{\tabcolsep}{1.2mm}
\begin{tabular}{ccc}
\toprule
\textbf{Majority Vote} & \textbf{Atleast One Vote} & \textbf{Krippendorff's Alpha}\\
\midrule
0.55 & 0.75 & 0.43 \\
\bottomrule
\end{tabular}
\caption{Human Agreement Study on the union of RESIN+CuratedSchemas (Zero/One-Shot). 75\% of examples judged equivalent by the entailment model are judged equivalent by at least one Turker. Turker ratings are in moderate agreement according to Krippendorff's $\alpha$.}
\label{tab:human-eval}
\end{table}

Overall, we see promising results on event schema drafting performance with language models with minimal human intervention and ability to automatically evaluate against gold schemas.


\subsection{Stylistic and Coverage Differences}
\label{sec:style}
In this section, we investigate the various differences that can occur in natural language schemas derived from different sources.

\paragraph{There are stylistic differences between event schema datasets and generations.} Our gold datasets are derived from two independent sources and have stylistic differences in the method of representing natural language events. 
In Table \ref{tab:word-length} we show the average length of these prompts. We see that the mean length of sentences measured by word count varies between 3.57 to 6.29 among the datasets and LM generated schemas. Some stylistic influence can be achieved by one-shot prompting as noted in the case of word count difference between zero-shot and one-shot outputs of \texttt{text-davinci-003}. We also show this qualitatively in Section~\ref{sec:qualitative}. 

\paragraph{One-shot prompts for style-matching with gold schemas} We can much better match the style of schemas by providing them as demonstrations in one-shot prompts. Specifically, for generating a schema for domain \textit{d}, we formulate one-shot prompts as shown in appendix \ref{sec:appendix:prompts} from three domains \textit{x}, $\forall x \notin [d]$. 

\paragraph{Inter-dataset Agreement} To further confirm that the schemas we have differ, Table \ref{tab:human-agreement} shows that the average event recall measured between the gold schemas of RESIN-11 and CuratedSchemas. This result conflates two things: the performance of the entailment model (discussed more in Section~\ref{sec:human_eval_entailment}) and the meaningful differences in events between the two schemas. However, on inspection, stylistic attributes are responsible for both, as certain more specific events in RESIN have no analogue in CuratedSchemas due to the different styles. Entailment reflects this even though it is not reliable on every case.

\section{Human Evaluation of Entailment}
\label{sec:human_eval_entailment}

Our recall values in Table~\ref{tab:event-recall-multi} are high enough to establish the utility of our approach. Most events can theoretically be matched to some other event in our generated dataset.

To confirm whether the entailment systems are making correct decisions, we conduct a precision-focused human evaluation of the automatic entailment decisions. The objective was to assess how reliably the entailment models aligned with our actual judgments regarding the equivalence between events. To gather annotations for this evaluation, we used Amazon Mechanical Turk (AMT) and enlisted the participation of randomly selected human annotators. We presented them with 216 sampled event pairs from all domains consisting of gold and predicted events that are matched by the any-directional entailment model as described in \ref{sec:entailment-eval}. The annotators were then asked to indicate their agreement with each match. Each task is annotated by three unique annotators to measure overall consensus. Further details on the task setup can be found in Appendix \ref{sec:anno-task-desc}.

The results of our human agreement study are shown in Table \ref{tab:human-eval}. The event match performance of entailment models across two datasets and all event domains achieves a majority vote agreement of 0.55 with the entailment judgments. However, at least one annotator agrees with the event match 75\% of the time. We also measure the Krippendorff’s Alpha to measure the inter-annotator agreement. The alpha score for our task is 0.43, which is considered moderate agreement, but does reflect the subjectivity of the task. 




We argue that not all of the entailment mistakes labeled as such truly represent errors. For instance, the any-directional entailment model matches the prediction ``\textit{Implementation of preventative measures}'' to two gold events: ``\textit{people maintain physical distancing to prevent disease spread}'' and ``\textit{people are vaccinated against the disease}.'' Although the level of specificity differs between the two, we argue that \emph{any-directional entailment} can be a reasonable candidate for automatic metric while serving the purpose of assigning soft matches between gold and predicted events. Cases like this are often marked as not equivalent by Turkers, but we argue that the entailment judgment is still a reliable method for assessing recall.

For a highly precise evaluation protocol, \emph{bi-directional entailment} can be a suitable candidate, however, as this is a very strict metric, the recall achieved by this evaluation protocol is significantly lower (see Table~\ref{tab:bi-directional} in the Appendix).

We also conduct a internal human evaluation of the entailment metric at a granular level in Appendix \ref{sec:appendix:human-agree}.


\paragraph{The performance of entailment depends on stylistic matching} 
Table \ref{tab:human-agreement} highlights that human agreement of anydirectional entailment improves across all domains for \texttt{davinci-003} when the schemas are generated with one-shot prompts compared to zero-shot. This signifies that one-shot prompts are beneficial in guiding the language models to generate schemas of a specific style.

\renewcommand{\arraystretch}{1}
\begin{table*}[h]
	\centering
	\footnotesize
    \centering
	\begin{tabular}{ccl}
		\toprule
		\multicolumn{1}{c}{\textbf{LM/Gold Dataset}} & \multicolumn{1}{c}{\textbf{Prompt}} & \multicolumn{1}{c}{\textbf{Example}}\\
		\midrule
		 \multirow{3}{*}{\texttt{RESIN-11}} & \multirow{3}{*}{-} & medical treatment is attempted on infected people  \\
		 & & people donate to help fight the disease outbreak \\ & & officials are assigned to monitor, prevent, contain, and mitigate the disease outbreak \\
            \midrule
        \multirow{3}{*}{\texttt{CuratedSchemas}} & \multirow{3}{*}{-} &  disease control agency investigates outbreak \\
		 & & infected group reports to disease control agency \\ & & scientists invent drug \\
            \midrule
          \multirow{3}{*}{\texttt{text-davinci-003}} & \multirow{3}{*}{zero-shot} & ongoing monitoring of the disease \\
		 & & collaboration between healthcare providers and public health agencies \\ & & stockpiling of necessary medical supplies. \\
            \midrule
          \multirow{3}{*}{\texttt{flan-t5-xxl}} & \multirow{3}{*}{zero-shot} & people living in that country become infected with the pathogen
  \\
		 & & vaccines are developed and distributed
 \\ & & the laboratories informs the public about the disease outbreak \\
            \midrule
          \multirow{3}{*}{\texttt{text-davinci-003}} & \multirow{3}{*}{one-shot} & medical teams conduct research on the disease  \\
		 & & affected area is monitored for further outbreaks \\ & & people are vaccinated against the virus \\
            \midrule
          \multirow{3}{*}{\texttt{flan-t5-xxl}} & \multirow{3}{*}{one-shot} & government issues a public health advisory  \\
		 & & people are quarantined \\ & & disease is transmitted from animal to human \\
		\bottomrule 
	\end{tabular}
\caption{Examples of output events from the different gold annotations we use and large language models.}
\label{tab:examples-all}
\end{table*}

\section{Qualitative Analysis}
\label{sec:qualitative}
While the length analysis in Table~\ref{tab:word-length} shows differences between various domains and schema sources, the stylistic differences go beyond length in ways that are hard to precisely quantify. We show examples from the  \textbf{disease-outbreak} domain in Table~\ref{tab:examples-all} to highlight these differences and qualitatively depict the variation in the writing style of events across human-curated datasets (\texttt{RESIN-11} and \texttt{CuratedSchemas}) and generations from language models (\texttt{text-davinci-001} and \texttt{flan-t5-xxl}) in zero-shot and one-shot settings. 

We see that event samples from \texttt{AltShemas} are more formal and shorter in size as compared to \texttt{RESIN-11} which have a high length variance and are more natural language like. This style also differs from the zero-shot generations from \texttt{davinci-003} and \texttt{flan-t5-xxl}. A controlled generation using one-shot prompts derived from \texttt{RESIN-11} schema can be used to attempt to match the event description style of the gold schemas.

\section{Related Work}

\paragraph{Event-centric modeling and schema induction} Methods performing schema induction can be categorized into simple and complex schema induction. Simple schema induction methods rely on identifying event triggers and participants and do no incorporate the relationships between events \citep{chambers2013event, cheung2013probabilistic, nguyen2015generative, sha2016joint, yuan2018open}. Recent work \citep{li-etal-2021-future,du2022resin} focuses on generating complex schemas that incorporate temporal as well as event argument relationships but assume availability of large amount of event relevant corpora. Existing event datasets such as MAVEN \citep{wang2020maven} and event-centric knowledge bases such as Event-Wiki \citep{ge2018eventwiki}, but working with these datasets naturally restricts a system designer to a fixed ontology.

Closest to our work, \citet{zhang2023humanintheloop} also generate schemas including a GPT prompting stage. However, they follow this with a stage of grounding to an ontology, sidestepping the challenges with evaluation we tackle in this work and losing the ability to homogenize between two different sources of schemas. \citet{dror2022zero} use language models to generate large number of source documents about a topic that can be used to extract events and relations to build schemas in a zero-shot manner. However, their method uses language models to generate documents containing relevant information which is further used to extract events using event extraction methods. In this work, we provide a way to both generate and automatically evaluate light event schemas in natural language making the process less dependent on traditional event extraction and schema matching pipleines.

\paragraph{Textual Entailment} Natural Language Inference research focuses on establishing entailment between a premise and a hypothesis pair. Although most of the previous work focuses on sentence level hypothesis and premise pair, recent datasets such as DocNLI \citep{yin2021docnli} and ContractNLI \citep{koreeda2021contractnli} push the boundaries to extend NLI models to longer  multi-sentence inputs and real-work datasets.  \citet{schuster2022stretching} explore the utility of NLI models on longer inputs using a ``stretching'' form of aggregation, namely maxing over possible alignments to a document. 

It is common to see similarity models being used to judge similarity between two sentences in ROUGE, BLEU and BERTScore \cite{zhang2020bertscore}. However, recent works recommend the usage of NLI models as evaluation metrics for Abstract Summarization \citep{maynez2020faithfulness} as they capture the faithfulness and factuality of summaries better than standard metrics. \citep{zhang2021finding} explore the usage of NLI models to automate evaluation of summarization tasks which can also benefit automated best model checkpointing. In this work, we explore using NLI as a metric for schema coverage matching directly in natural language. 

\section{Conclusion}

In this paper, we explored the ability of language models to draft light event schemas. In both zero- and one-shot settings, we showed that large language models can generate coherent, varied sequences of events in natural language that overlap substantially with human-curated events across several domains of interest. We show that textual entailment can be used to evaluate these matches. We believe our work can pave the way for future efforts looking at how explicit knowledge like schemas can be used in tandem with large language models to make predictions. Streamlining the ability to generate schemas and then curate them with human intervention will be an important step to scaling this method to work across many domains.

\section*{Limitations}

The schemas we produce in this work are, by choice, lighter weight than representations used in some prior work. Past work \cite{li-etal-2021-future} has explored schemas with graph-structured ordering. While these schemas can express a richer set of partial ordering and mutual exclusion relationships between events, they are both cumbersome to produce and relatively little work has shown the ability to use them to perform complex inferences. Our view is that more complex structural relationships should also be specified in natural language for maximal compatibility with prediction based on large language models; we use this for future work. Human curation can also be used to impart these features for use in downstream applications.

A second limitation is that the robustness of event recall evaluation using textual entailment is dependent on the stylistic similarities between generated and gold schemas.  While we analyze this in the paper, stronger textual entailment systems down the road can potentially be useful to improve the precision of our performance estimates further.

Finally, we note that schema-mediated prediction with neural models is an emerging and ongoing area of research. Therefore, there are not standard systems we can plug our schemas into for downstream evaluation. Nevertheless, we believe that these knowledge structures can be intrinsically evaluated, and high quality representations will pave the way for future work in this area.



\bibliography{anthology,custom}
\bibliographystyle{acl_natbib}

\appendix
\section{Prompts}
\label{sec:appendix:prompts}
\paragraph{Zero-shot Prompts} The \texttt{text-davinci-003} zero-shot experiments predominantly use the prompt below. The domains \textit{d} that we consider are \texttt{natural disaster, disease outbreak, international conflict, mass shooting, IED attack, and kidnapping}

\subparagraph{Temporally-aided prompt}
\begin{quote}
    "List 10 things that each happen (1) before; (2) during; and (3) after a [d]? 

Before a [d], there are several things that can happen: 

1."
\end{quote}

\paragraph{Prompt Union} This approach uses three different prompt templates as shown below. In addition to the previously described prompt above, the two additional prompts that we sample for prompt union experiments are:

\subparagraph{Causes prompt}
\begin{quote}
    "List causes and events that can happen over the course of a [d]?

     Causes of a [d]:

     1."
\end{quote}

\subparagraph{Causes and temporally-aided prompt}

\begin{figure*}
    \centering
    \includegraphics[scale=0.70]{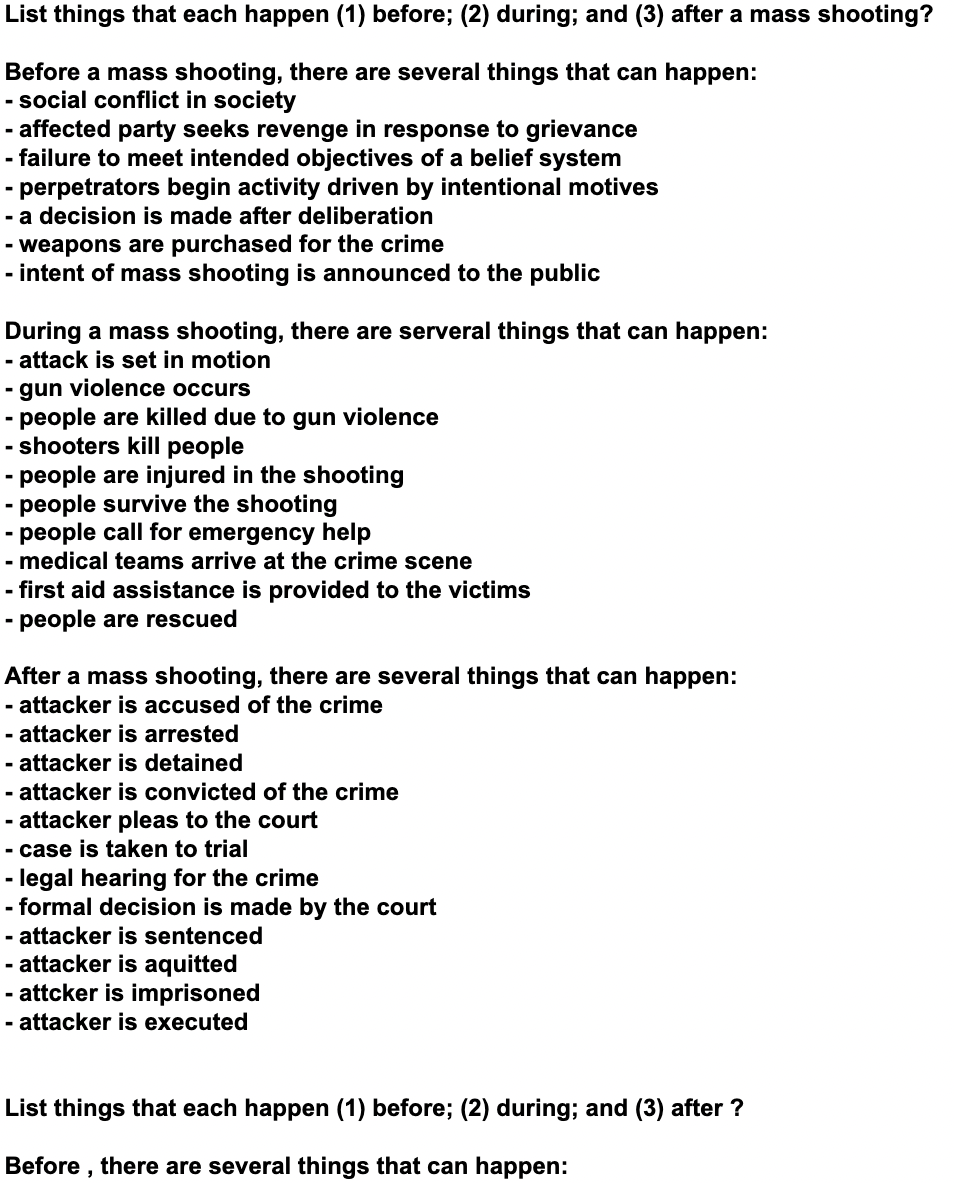}
    \caption{Example of one-shot prompt used to generate events.}
    \label{fig:one-shot}
\end{figure*}

\begin{quote}
    "List causes and events that can happen before, during and after a [d]?

    Causes of a [d]:
        
    1."
    
\end{quote}
\paragraph{One-Shot Prompts} 
Figure \ref{fig:one-shot} depicts a sample of the one-shot prompts we use to generate events. For generating events for one domain we sample from such prompts from three other domains and generate the output.

\section{Annotation Task Description} 
\label{sec:anno-task-desc}
\begin{figure*}
    \centering
    \includegraphics[scale=0.75]{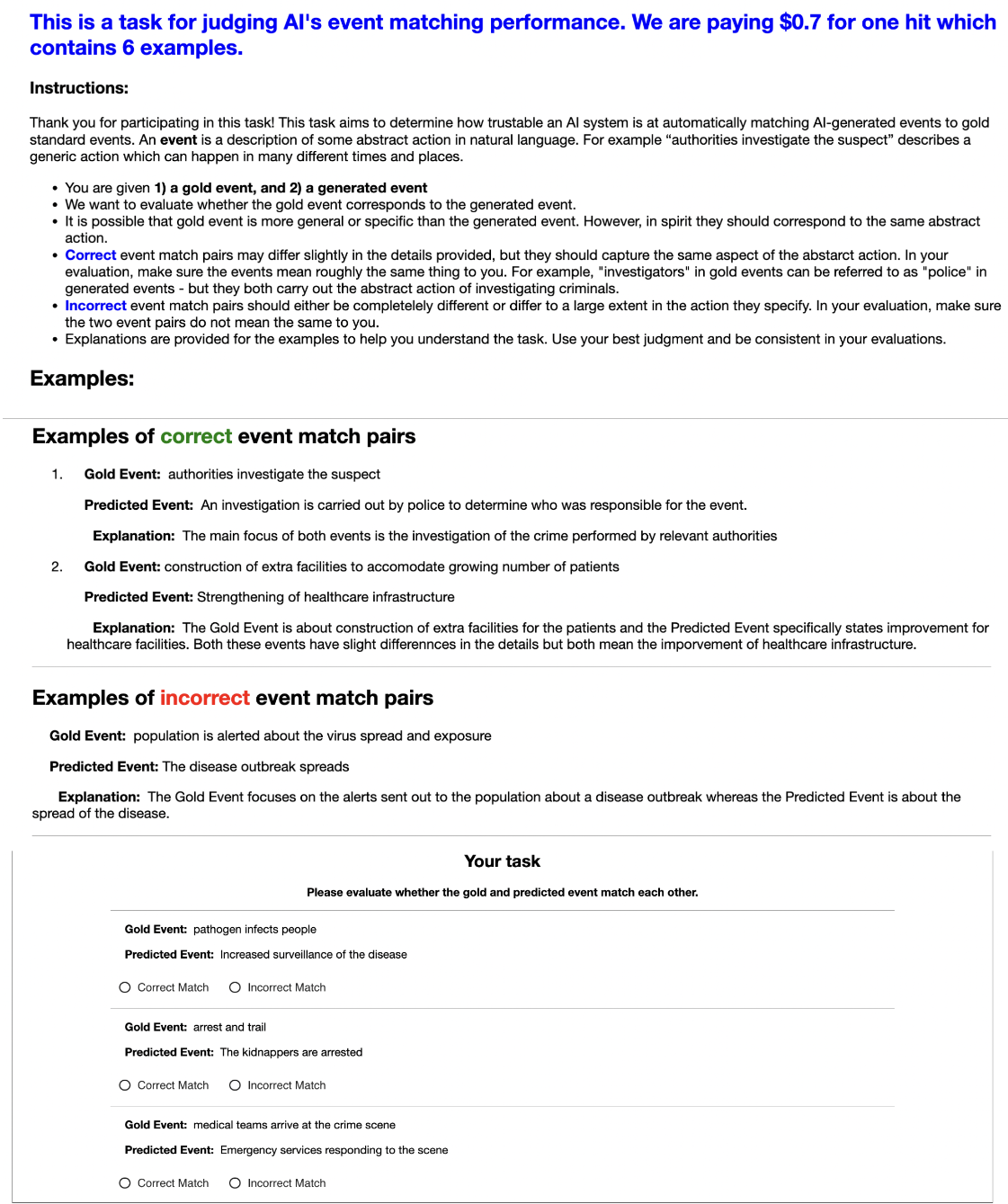}
    \caption{Task Description for Human Agreement Annotation for Event Matching on AMT}
    \label{fig:amt_task}
\end{figure*}

The task description provided to Turk workers on Amazon Mechanical Turk is shown in Figure \ref{fig:amt_task}. 

\section{Comparison of any-directional and bi-directional entailment over event recall} We compare the event recall values predicted by anydirectional and bidirectional entailment as a metric in \ref{tab:bi-directional}.

\begin{table*}
\centering
\small
\begin{tabular}{llll}
\toprule
\textbf{Domain} & \textbf{Gold Schema} & \textbf{Any-directional} & \textbf{Bi-directional}\\
\midrule
Disease Outbreak & RESIN & 0.38 & 0.05\\
 & CuratedSchemas & 0.29 & 0.06  \\
\midrule
Kidnapping & RESIN & 0.42 & 0.05 \\
 & CuratedSchemas & 0.54 & 0.02 \\
 \midrule
 International Conflict & RESIN & 0.46 & 0.07 \\
 & CuratedSchemas & 0.55 & 0.11 \\

 \midrule
 Natural Disaster & RESIN & 0.56 & 0.06 \\
 & CuratedSchemas & 0.4 & 0.06 \\
 \midrule
 IED & RESIN & 0.53 & 0.08 \\
 & CuratedSchemas & 0.42 & 0.09 \\
\bottomrule
\end{tabular}
\caption{Event recall difference between anydirectional and bidirectional entailment. }
\label{tab:bi-directional}
\end{table*}

\section{Example of \texttt{davinci-base} output}
\label{sec:appendix:davinci-base}
\begin{figure*}
    \centering
    \includegraphics[scale=0.70]{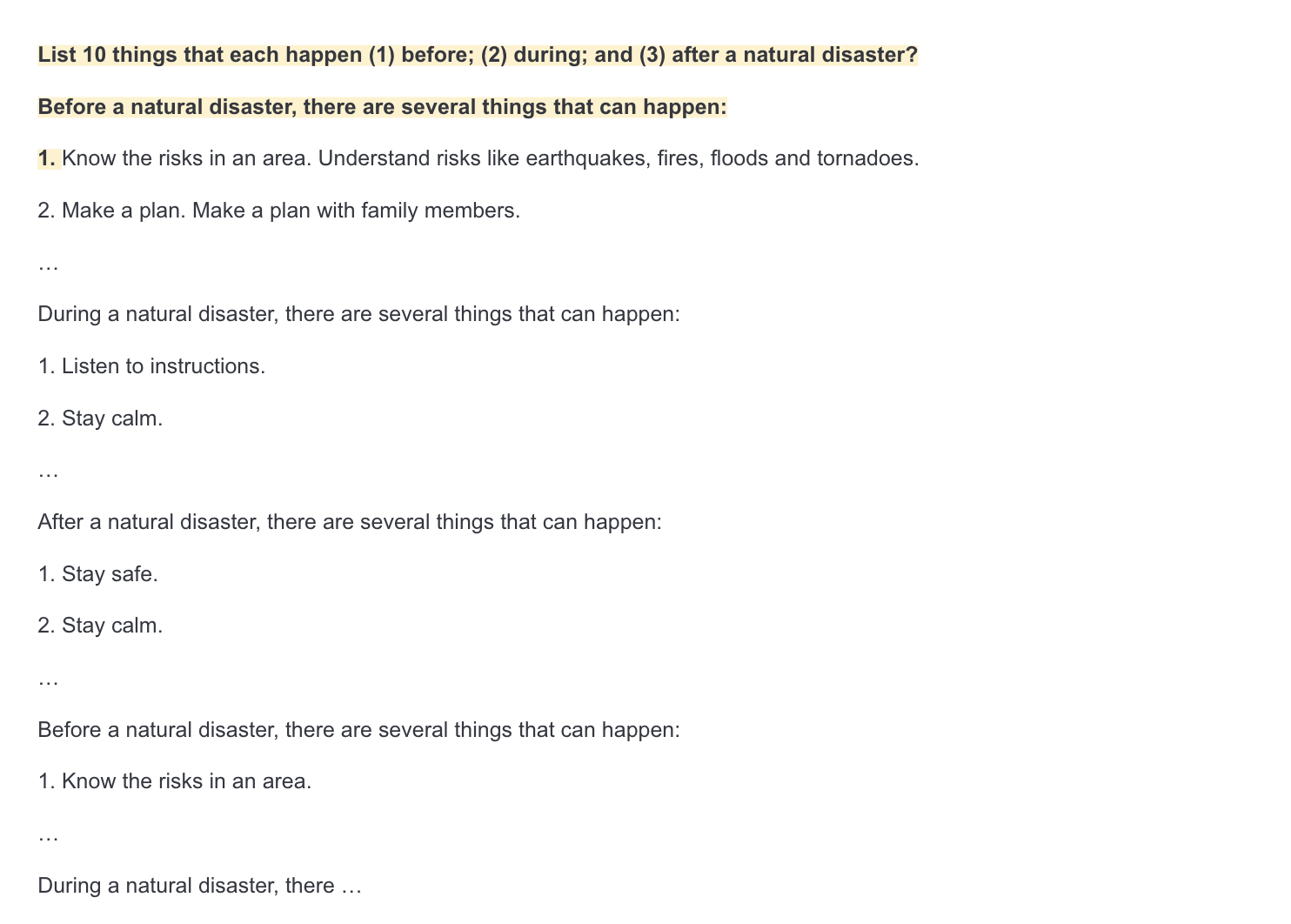}
    \caption{Examples of output from davinci-base.}
    \label{fig:davinci-base}
\end{figure*}

Figure \ref{fig:davinci-base} shows a representative example for output from davinic-base using zero-shot prompt.

\begin{table*}
\centering
\small
\begin{tabular}{llll}
\toprule
\textbf{Domain} & \textbf{Gold Label} & \textbf{Prediction} & \textbf{Human Agreement}\\
\midrule
IED Attack & RESIN & CuratedSchemas & 0.74 \\
IED Attack & CuratedSchemas & RESIN &  0.77 \\
 \midrule
Disease Outbreak & RESIN & CuratedSchemas & 0.38 \\
Disease Outbreak & CuratedSchemas & RESIN & 0.8 \\
\midrule
All Domains & RESIN+CuratedSchemas & \texttt{davinci-003} (zero-shot ) & 0.7 \\
All Domains & RESIN+CuratedSchemas & \texttt{flan-t5-xxl} (zero-shot) &  0.61\\
All Domains & RESIN+CuratedSchemas & \texttt{davinci-003} (one-shot) & 0.79 \\
\bottomrule
\end{tabular}

\caption{
Granular Human Agreement Study 
}
\label{tab:human-agreement}
\end{table*}

\begin{figure*}
    \centering
    \includegraphics[scale=0.50]{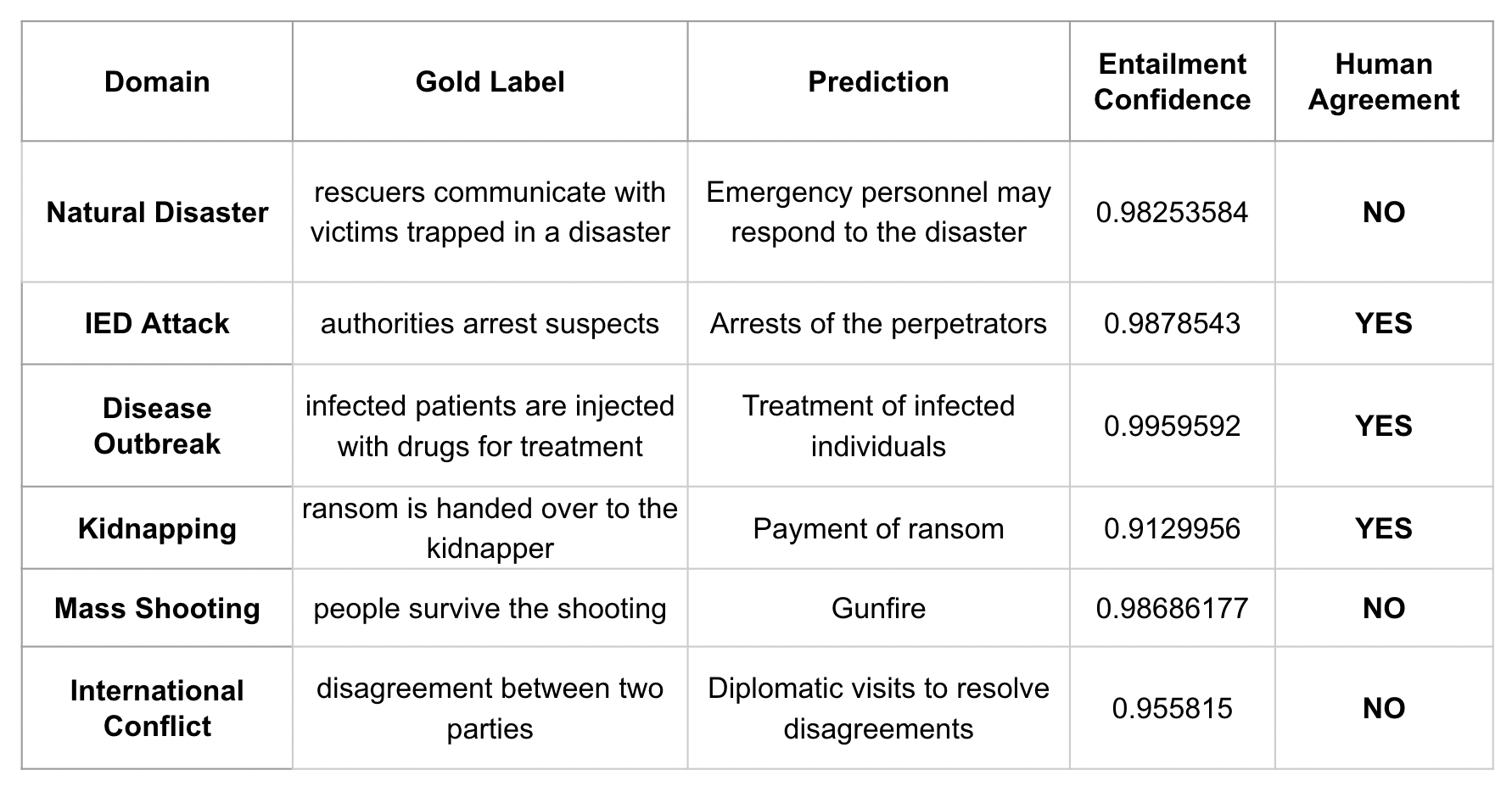}
    \caption{Examples of annotation for human agreement with the predictions of the any-directional entailment model.}
    \label{fig:human-agee}
\end{figure*}

\section{Dataset Description}
\label{sec:dataset-stats}
In this work we experiment with a subset of two datasets: 1) RESIN-11 and 2) CuratedSchemas. The dataset statistics of the subsets are described in Table \ref{tab:dataset-stats}.
\begin{table*}
\begin{tabular}{llll}
\toprule
\textbf{Dataset} & \textbf{Domain Count} & \textbf{Average No. of Events} & \textbf{Avg. Length of Events}\\
\midrule
RESIN-11 & 6 & 34.33 & 5.75\\
CuratedSchemas & 6 & 38.33 &  3.57\\
\bottomrule
\end{tabular}
\caption{Dataset Statistics for RESIN-11 and CuratedSchemas used in this work.}
\label{tab:dataset-stats}
\end{table*}

\begin{figure*}
    \centering
    \includegraphics[scale=0.5]{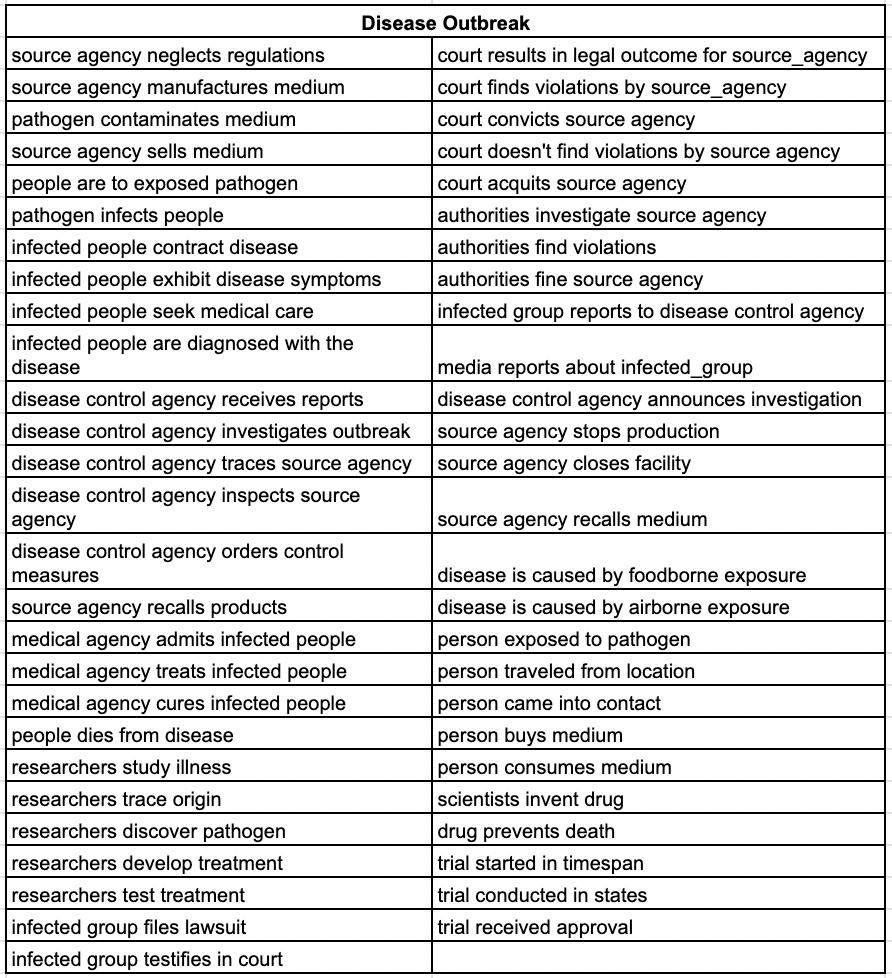}
    \caption{Example of events in the Disease Outbreak Schema from CuratedSchemas dataset}
    \label{fig:curated-schemas}
\end{figure*}

\subsection{CuratedSchemas}
\label{sec:curated_schemas}

\begin{figure*}
    \centering
    \includegraphics[scale=0.5]{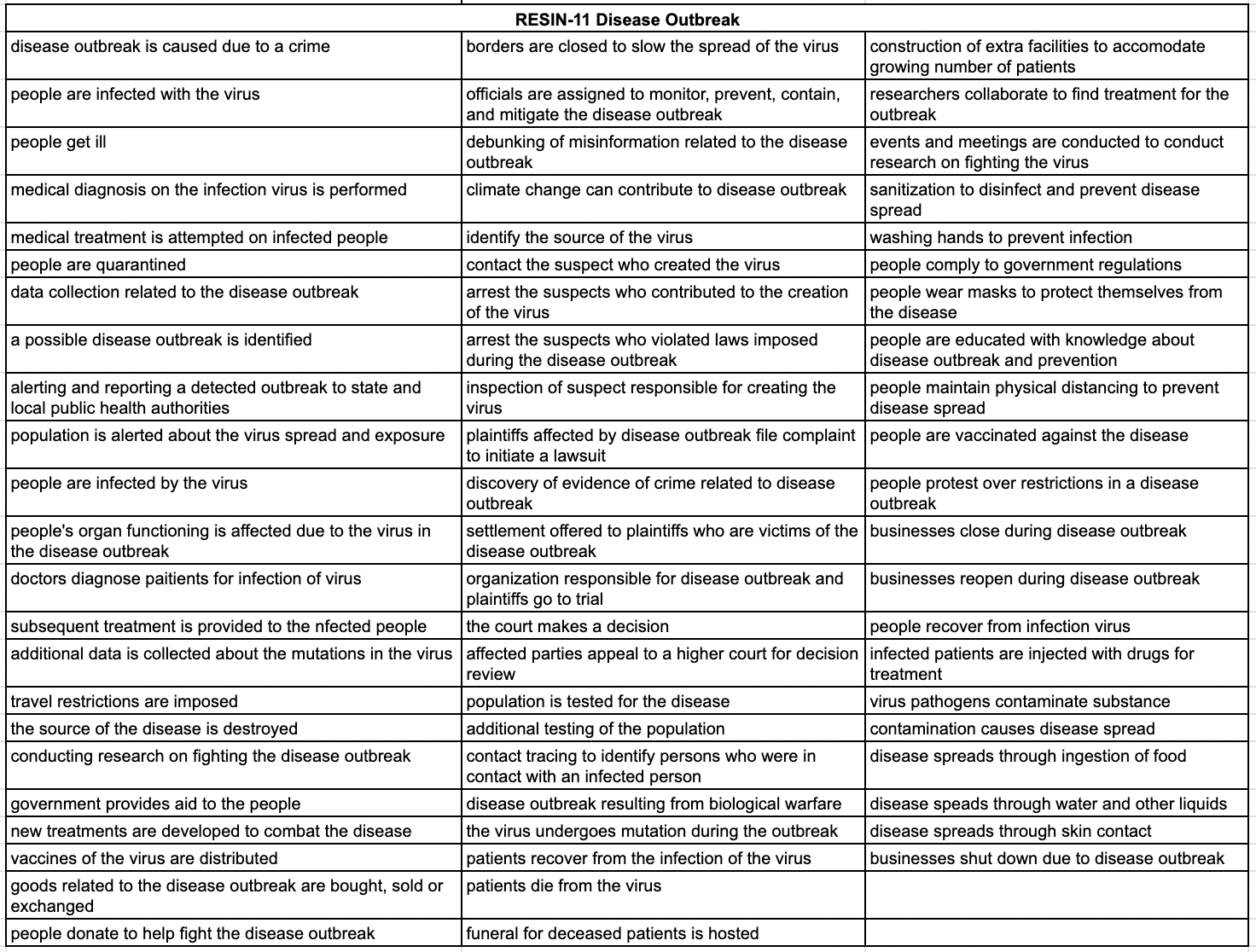}
    \caption{Example of events in the Disease Outbreak Schema derived from Resin-11 dataset}
    \label{fig:resin-schemas}
\end{figure*}

CuratedSchemas was produced over a period of roughly 1 year. Around 10 distinct annotators, all experts in natural language processing, contributed events to build up these schemas over several stages. Multiple rounds of feedback were collected based on a downstream prediction task and human assessment of outputs from that task, not conducted by the authors of this paper. The schemas balance precision and recall, aiming to cover each domain thoroughly but not include events that are too low-level or would never conceivably be reported in text. Additional layers of structure are presented in the CuratedSchemas dataset, but here we only evaluate against the collection of abstract events. An example of events from the Disease Outbreak domain of CuratedSchemas is shown in Figure \ref{fig:curated-schemas}.

\subsection{RESIN Schemas}
\label{sec:resin_schemas}

The natural language description of the event is derived from the published RESIN schemas using the fields of \texttt{qnode}, \texttt{qlabel}, \texttt{description}, and/or \texttt{ta1explanation} as applicable. An example of the events from the Disease Outbreak domain derived from schemas published in RESIN-11 is shown in Figure \ref{fig:resin-schemas}.

\section{Granular Human Agreement Study}
\label{sec:appendix:human-agree}
We perform a granular human agreement study to investigate the trends and impacts that various domains, prompting methods, and language models have on the performance of entailment models for automatic evaluation. This study is performed by the authors of the paper and is illustrated in Table \ref{tab:human-agreement}.

Figure \ref{fig:human-agee} shows examples of our annotation across different domains for human agreement with the pairings predicted by the any-directional entailment model between gold and generated schema events.

\end{document}